\documentclass[10pt,twocolumn,letterpaper]{article}

\usepackage[pagenumbers]{cvpr} 

\usepackage{graphicx}
\usepackage{amsmath}
\usepackage{amssymb}
\usepackage{booktabs}

\usepackage{color}
\usepackage{multirow}
\usepackage{float}

\usepackage[pagebackref,breaklinks,colorlinks]{hyperref}

\definecolor{pp}{rgb}{0.6,0.0,0.6}

\usepackage[capitalize]{cleveref}
\crefname{section}{Sec.}{Secs.}
\Crefname{section}{Section}{Sections}
\Crefname{table}{Table}{Tables}
\crefname{table}{Tab.}{Tabs.}

\def\cvprPaperID{11457} 
\def\confName{CVPR}
\def\confYear{2022}

\usepackage{overpic}
\usepackage{enumitem} 
\usepackage{overpic} 
\usepackage{color}

\definecolor{turquoise}{cmyk}{0.65,0,0.1,0.3}
\definecolor{purple}{rgb}{0.65,0,0.65}
\definecolor{dark_green}{rgb}{0, 0.5, 0}
\definecolor{orange}{rgb}{0.8, 0.6, 0.2}
\definecolor{red}{rgb}{0.8, 0.2, 0.2}
\definecolor{darkred}{rgb}{0.6, 0.1, 0.05}
\definecolor{blueish}{rgb}{0.0, 0.3, .6}
\definecolor{light_gray}{rgb}{0.7, 0.7, .7}
\definecolor{pink}{rgb}{1, 0, 1}
\definecolor{greyblue}{rgb}{0.25, 0.25, 1}

\newcommand{\Figure}[1]{Figure~\ref{fig:#1}}

\newcommand{\Table}[1]{Table~\ref{tab:#1}}

\newcommand{\Equation}[1]{Equation~\ref{eq:#1}}

\usepackage{blindtext}

\renewcommand{\paragraph}[1]{\vspace{1em}\noindent\textbf{#1}.}

\newcommand{\methodname}[0]{DEER}

\begin{document}
\title{DEER: Detection-agnostic End-to-End Recognizer for Scene Text Spotting}

\author{Seonghyeon Kim\thanks{Correspondence to \texttt{kim.seonghyeon@navercorp.com}}\quad Seung Shin\quad Yoonsik Kim \quad Han-Cheol Cho\quad Taeho Kil \\ \quad Jaeheung Surh\quad Seunghyun Park\quad Bado Lee\quad Youngmin Baek \\
\vspace{1.3ex}
NAVER Clova 
}
\maketitle

\begin{abstract}

Recent end-to-end scene text spotters have achieved great improvement in recognizing arbitrary-shaped text instances. Common approaches for text spotting use region of interest pooling or segmentation masks to restrict features to single text instances. However, this makes it hard for the recognizer to decode correct sequences when the detection is not accurate \ie one or more characters are cropped out. Considering that it is hard to accurately decide word boundaries with only the detector, we propose a novel Detection-agnostic End-to-End Recognizer, \methodname, framework. The proposed method reduces the tight dependency between detection and recognition modules by bridging them with a single reference point for each text instance, instead of using detected regions. The proposed method allows the decoder to recognize the texts that are indicated by the reference point, with features from the whole image. Since only a single point is required to recognize the text, the proposed method enables text spotting without an arbitrarily-shaped detector or bounding polygon annotations. Experimental results present that the proposed method achieves competitive results on regular and arbitrarily-shaped text spotting benchmarks. Further analysis shows that \methodname~is robust to the detection errors. The code and dataset will be publicly available.

\end{abstract}

\section{Introduction}

End-to-end scene text spotting has recently gained a lot of attention due to its simplicity and performance gain. These works also have many practical applications in various areas, such as information extraction, image retrieval, and visual question answering. Commonly, an end-to-end text spotting pipeline consists of a text detector and a recognizer. The detector outputs a box or polygon representation to localize the text instances within an image, and the recognizer takes each localized text region as an input to decode the characters within each patch of the image.




\begin{figure}[t]
  \centering
  \begin{subfigure}{0.98\linewidth}
    \includegraphics[width=1.0\linewidth]{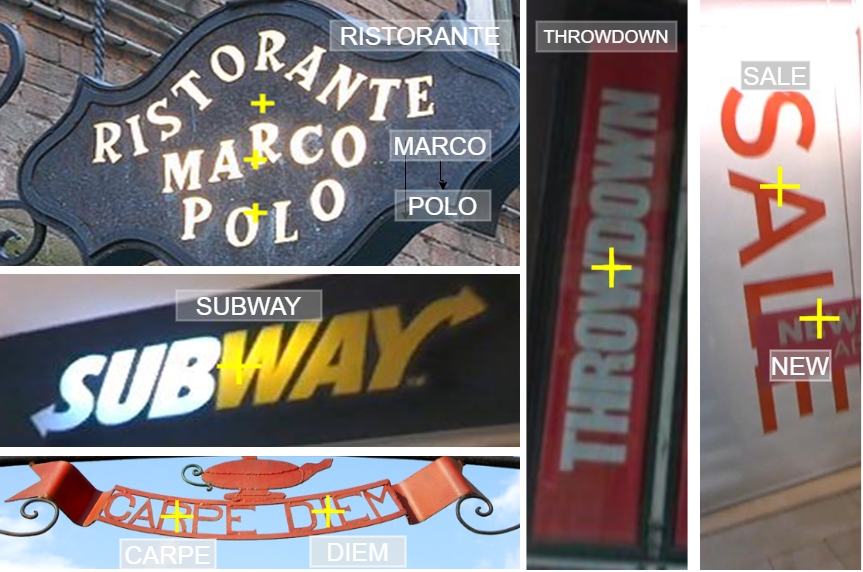}
    \caption{Outputs of \methodname~on challenging examples.}
    \label{fig:teaser_successful}
  \end{subfigure}
   \begin{subfigure}{0.46\linewidth}
    \includegraphics[width=1.0\linewidth]{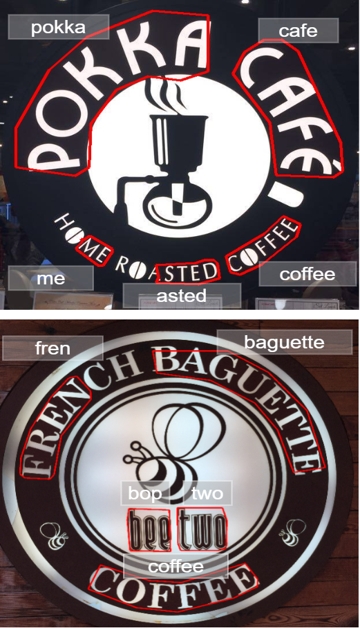}
    \caption{Outputs of a segmentation based model with some incorrect predictions.}
    \label{fig:teaser_mts}
  \end{subfigure}
   \begin{subfigure}{0.46\linewidth}
    \includegraphics[width=1.0\linewidth]{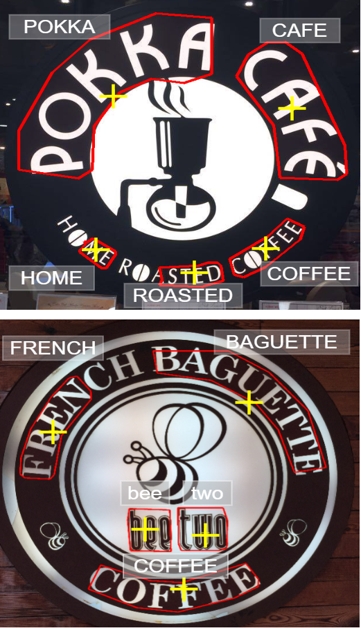}
    \caption{Outputs of \methodname~when the reference point is obtained from (b).}
    \label{fig:teaser_mts_deer}
  \end{subfigure}
   \caption{(a) is the successful outputs of \methodname~on challenging samples from other papers~\cite{baek2020character,liao2020mask}. + indicates the sampled reference point. (b) is the outputs of a comparison model with some incorrect predictions. The red outline indicates detection output. (c) is the output of \methodname~when the reference point is obtained from (b)'s detection output that involves incorrect detection.}
    \label{fig:teaser}
\end{figure}

\begin{figure*}[t]
\begin{center}
\includegraphics[width=1.0\linewidth]{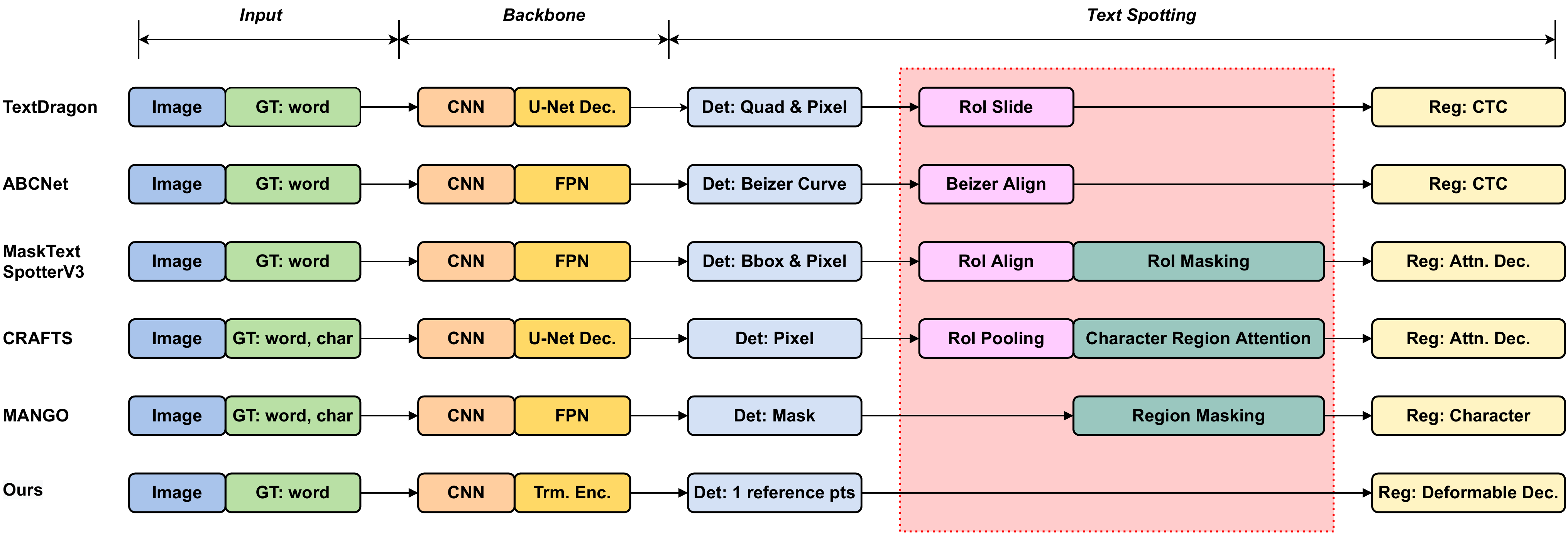}
\end{center}
   \caption{An overview of various end-to-end text spotting methods. The proposed method is different from other alternatives in that it does not use any pooling or masking technique.}
\label{fig:related_works}
\end{figure*}

Previous scene text spotting pipelines used a rather tightly coupled framework between detector~\cite{CUTE80,liao2018textboxes++,baek2019character,liao2020real} and recognizer~\cite{Jaderberg16mjsynth,shi2016robust,FAN,CRNN,AON,ASTER,ESIR}. Specifically, cropped $images$ from the detector are fed to the recognizer, which inevitably means that the recognition performance is critically dependent on that of the detector and image patches it outputs. 
Recently, end-to-end text spotting methods~\cite{liu2018fots,baek2020character,liao2020mask,liu2020abcnet} proposed the use of a more loosely coupled framework by utilizing ROI pooling or masking to extract $features$ and restrict the input region for the recognizer to contain a single word.
Although taking the localized features for the recognizer could lessen the dependency of the recognizer on the cropped regions from detectors, the errors from the detector are still accumulated, and thus, these errors can cause recognition failure cases as shown in \Figure{teaser_mts}. 
Moreover, feature pooling and masking still require data with bounding boxes (or bounding polygons) to train end-to-end text spotting models, even if the final application does not require exact bounding box information.

With the advancement of recent end-to-end Transformer-based approaches in the field of object detection~\cite{carion2020end,zhu2021deformable,yao2021efficient,chen2021pix2seq}, it is becoming more evident that exact region information, sophisticated ground truth assignment, and feature pooling are not strictly required to recognize individual objects in the image.
From these observations, we propose a novel Detection-agnostic End-to-End Recognizer, denoted as \methodname, that drastically relieves the dependency on the exactness of detection results. 
Instead of relying on detectors to extract accurate text regions, we let the detector localize a single reference $point$ for each text instance. Then, we design a text decoder to comprehensively recognize texts around the corresponding reference point. 
Specifically, given a reference point, the text decoder learns to determine the attending region of that specific text instance while decoding the text sequence.  
Since \methodname~only requires the detector to localize a single reference point, it allows the model to use a much wider variety of detection algorithms and annotations. Moreover, this approach can naturally handle rotated and curved text instances without pooling operations and polygon-type annotations.
As shown in \Figure{teaser_successful}, \methodname~successfully recognizes challenging samples mentioned in ~\cite{baek2020character,liao2020mask}. The samples present rotated text, text-in-text, and text with granularity issues. Thus, \methodname~achieves comparable performance to that of the state-of-the-art models. Moreover, we validate the detection agnostic property of \methodname~by investigating diverse ablation studies. \Figure{teaser_mts_deer} shows detection agnostic property of \methodname, where the reference point from other detection output is used together with the decoder in \methodname. Although the detected region is not accurate and the reference point is also biased, DEER correctly recognizes the words.  

\section{Related Works}
\begin{figure*}[t]
\begin{center}
\includegraphics[width=1.0\linewidth]{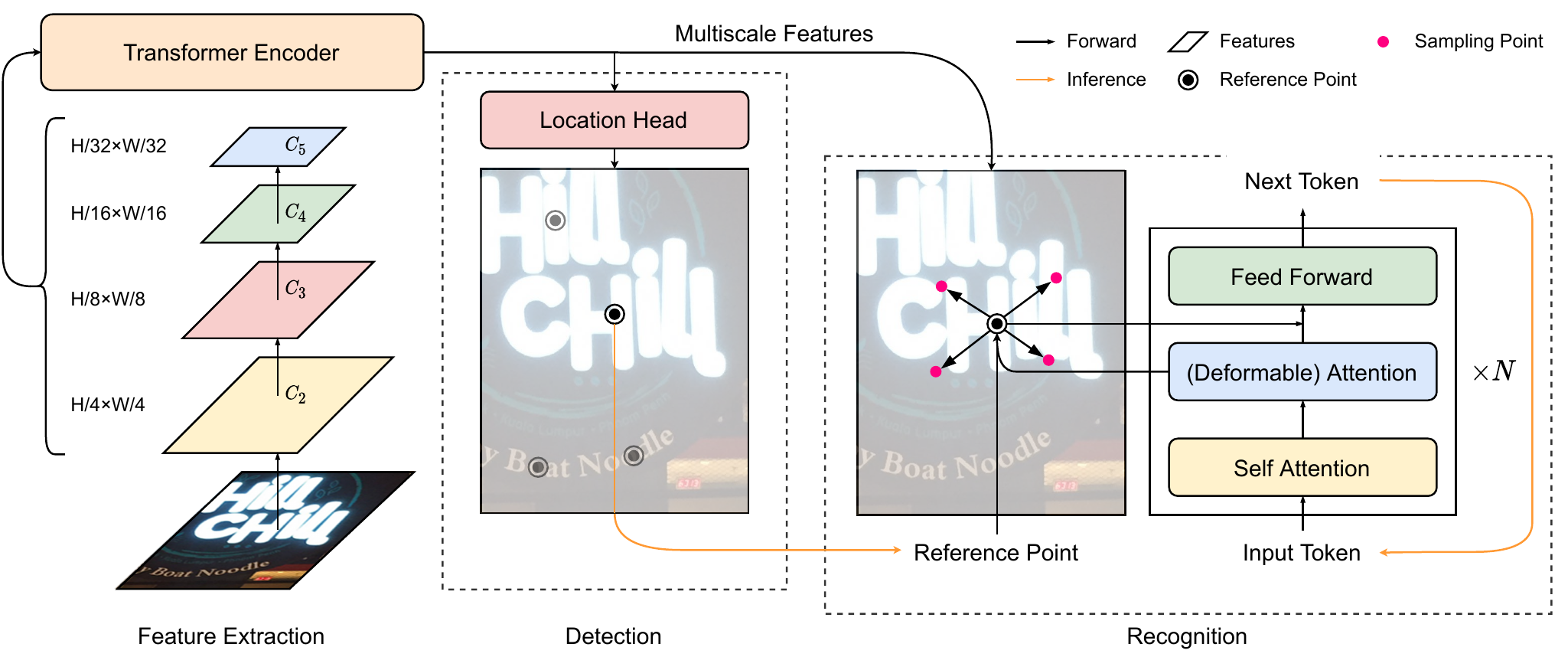}
\end{center}
   \caption{Overview of the proposed method. Refined feature tokens from the encoder is used as an input to the location head and the text decoder. During inference, center points from the location head is used as reference points.}
\label{fig:method}
\end{figure*}

In this section, we review various end-to-end text spotting models and study the latest object detection and segmentation models that inspired the development of the proposed method.



\subsection{End-to-end Scene Text Spotting}
\label{ss:e2e_sts}

Early end-to-end text spotting models, like TextBoxes++~\cite{liao2018textboxes++} for instance, separately trained the text detector~\cite{liao2017textboxes} and recognizer~\cite{CRNN}, and joined them afterwards. Following studies~\cite{liu2018fots,baek2020character,liao2020mask,liu2020abcnet} found that jointly training both detector and recognizer improved the final performance. Because the performance of these end-to-end models heavily relies on detection results, many works focused on developing sophisticated detection and feature pooling/masking algorithms. For example, ABCNet~\cite{liu2020abcnet} localized arbitrary-shaped text instances by using a Bezier-curve fitting algorithm. TextDragon~\cite{feng2019textdragon} used an RoI sliding algorithm to group a series of local features along the centerline. Another line of work utilized segmentation as its technique. For instance, MaskTextSpotterV3~\cite{liao2020mask} adopted hard RoI masking to remove features not related to the target text instances. MANGO~\cite{qiao2020mango} used an image-level mask attention, which doesn't need RoI operation.



As shown in \Figure{related_works}, while existing methods include additional modules to restrict the input features, the proposed method performs text spotting without these modules.
The proposed method is different from other end-to-end approaches in that the detector provides only a reference point to the recognizer. The recognizer exploits any information from the reference point and the entire input to decode the output text sequence.

\vspace{2mm}



\subsection{End-to-end Object Detection and Segmentation}
\label{ss:e2e_od}

Recent Transformer-based object detection and instance segmentation models have achieved impressive results.
DETR~\cite{carion2020end} showed competitive results without using sophisticated hand-crafted components, such as spatial anchor-boxes or non-maximum suppression.
Deformable DETR~\cite{zhu2021deformable} improved the training speed of DETR by employing deformable attention and solved the issue of performance on small objects by using multi-scale features.
Efficient DETR~\cite{yao2021efficient} further accelerated the training speed by using well-initialized reference points and object queries generated from the initial dense object detection stage.
For panoptic segmentation, Panoptic SegFormer~\cite{li2021panoptic} achieved state-of-the-art performance by using two-stage decoders, namely a location decoder and a mask decoder.

These studies showed that an explicit detection proposal is not required to achieve high recognition performance.
Based on this observation, we adopt the Transformer architecture and the concept of reference points (or location queries) for the end-to-end text spotting task, relaxing the dependency between detector and recognizer.

\section{DEER}




The overall pipeline of \methodname~is shown in \Figure{method}. \methodname~consists of a backbone, Transformer encoder, location head, and text decoder. First, the Transformer encoder combines the multi-scale feature maps generated by the backbone. Then, the location head predicts the reference points of text instances and (optionally) bounding boxes. Finally, the text decoder generates the character sequences in each text instance specified by their reference points. 

In the forward phase, an input image \(X\in\mathbb{R}^{H\times W\times 3}\) is fed into the backbone, and the feature maps \(C_2\), \(C_3\), \(C_4\), \(C_5\) are extracted. The resolution of each extracted feature corresponds to \(1/4\), \(1/8\), \(1/16\), \(1/32\), respectively. We then apply a fully-connected layer and group normalization~\cite{wu2018gn} on these feature maps to project into 256 channels. Then, we flatten and concatenate them into feature tokens with size \((L_2 + L_3 + L_4 + L_5)\times 256\), where \(L_i\) corresponds to the flattened length of \(C_i\), which is \(\frac{H}{2^i}\times\frac{W}{2^i}\). Then, using this as its input, the Transformer encoder outputs the refined features. We use features of size \(L_2\), which corresponds to \(C_2\) as the input to the location head. Finally, by using the reference points and the refined feature outputs, the text decoder generates the character sequences within the text instance autoregressively.



\subsection{Transformer Encoder}

Using high-resolution, multi-scale features is beneficial for text recognition. Since the computation cost of self-attention increases quadratically with input lengths, it is impractical to use Transformer on the concatenation of multi-scale features.  Therefore, previous approaches that employ Transformer encoders for object detection have used lower-resolution features like \(C_5\).

Unlike these previous methods, we use deformable attention, which scales linearly with input lengths. Due to the efficiency of deformable attention, our encoder is able to refine high-resolution concatenated features to generate multi-scale feature tokens \(F\).

\noindent\textbf{Deformable Attention} Deformable attention~\cite{zhu2021deformable} is a crucial component for both the encoder and decoder, due to its efficiency and location-awareness. Deformable attention is calculate by
\begin{multline}
\operatorname{DeformAttn}_h(A_{hqk}, p_\text{ref}, \Delta p_{hqk}) \\
= W_h^o\left[\sum_{k=1}^K A_{hqk}\cdot W_h^k x(v, p_{\text{ref}} + \Delta p_{hqk})\right],
\end{multline}
where \(x(v, p)\) is a bilinear interpolation that extracts features from value features \(v\) at position \(p\). \(K\) is the number of sampled key points and \(k\) is the key index. \(h\) is the index for the attention head, and \(W_h^o \in \mathbb{R}^{C \times C_m}\), \(W_h^k \in \mathbb{R}^{C_m \times C}\) are linear projections. \(p_{\text{ref}}\), \(\Delta p_{hqk}\), and \(A_{hqk}\) correspond to reference points, sampling offsets, and attention weights, respectively. \(p_{\text{ref}}\), \(\Delta p_{hqk}\), and \(A_{hqk}\) are computed by applying linear projections to the query features, and a softmax is applied on \(A_{hqk}\). In the encoder, we use fixed reference points with normalized coordinates on \([0, 1]\times[0, 1]\).

Instead of using predefined reference points, as shown in~\cite{yao2021efficient}, image-specific reference points can be used to accelerate model training. Also, the reference points positioned in each object can be used to decode specific object instances. We have used this feature to decode specific text instances.

\subsection{Location Head}

Utilizing location information is beneficial for recognizing and distinguishing objects. Inspired by previous approaches~\cite{yao2021efficient, li2021panoptic}, we use a location head to predict reference points (\ie, center position of text instance) for the text decoder.
Additionally, it provides a segmentation map to extract the bounding polygon of text instances, which is required only for computing evaluation metrics for our purposes. We adopted differentiable binarization (DB) \cite{liao2020real} to extract bounding polygon of text instances, which is required for the evaluation.

Specifically, we extract feature tokens of size \(L_2\) from \(F\), which corresponds to \(C_2\), then reshaped them to \((H/4, W/4)\). Then, to obtain binary and threshold maps we apply a separated segmentation head which consists of transposed convolution, group normalization, and relu. In the inference phase, the center coordinates of text instances are obtained from the detected bounding polygons and are used as the reference points.




\subsection{Text Decoder}

The recognition branch, which consists of a Transformer decoder, predicts the character sequences within the text instance autoregressively. The query \(Q\) for the text decoder is composed of the character embedding, positional embedding, and reference point \(q_{\text{ref}}\). The keys \(K\) and values \(V\) of the text decoder are the feature tokens \(F\) from the Transformer encoder. We pass the queries through self-attention, deformable attention, and feed-forward layers. Inspired by~\cite{zhu2019, li2021panoptic}, we also add a regular cross attention instead of deformable attention to the \(F\) alternatingly.


During training, \(N_t\) text boxes are sampled from the image and the computed center coordinates are used as reference points for the decoder. This allows for the independent training of the location head and text decoder. In the training stage, we use the points computed by the ground truth regions as reference points for the text decoder. For evaluation, the center coordinate from the detection branch is used as the reference point. To reduce the gaps between coordinates from the ground truth (training) and model prediction (evaluation), the center coordinates are perturbed in the training stage using the equation below:

\begin{equation}
\begin{gathered}[b]
q_{\text{ref}} = p_c + \frac{\eta}{2} \operatorname{min}(\|p_{\text{tl}} - p_{\text{tr}}\|, \|p_{\text{tl}} - p_{\text{bl}}\|), \\
\eta ~ \sim \operatorname{Uniform}(-1, 1)
\end{gathered}
\label{eq:perturb}
\end{equation}
where \(p_c\) is the centroid of the ground truth polygon, and \(p_{\text{tl}}\), \(p_{\text{tr}}\), \(p_{\text{bl}}\) correspond to the coordinates of the top-left point and adjacent top-right and bottom-left points. In the inference phase, the center point of the text regions extracted from the detection stage is used as the reference point.

\subsection{Optimization}

The loss function \(L\) used for training is defined as below:
\begin{equation}
L = L_{r}  + \lambda_s L_s + \lambda_b L_{b} + \lambda_t L_{t},
\end{equation}
where \(L_r\) is the autoregressive text recognition loss, and \(L_s\), \(L_b\), \(L_t\) are the losses from differentiable binarization~\cite{liao2020real}. Each denotes a loss for the probability map, binary map, and threshold map. \(L_r\) is computed with a softmax cross entropy between the predicted probability of character sequences and the ground truth text labels of corresponding text box. Following differentiable binarization, we apply a binary cross entropy with hard negative minining for \(L_s\), dice loss for \(L_b\), and \(L_1\) distance loss for \(L_t\).









During inference, we only use the probability map from the location head. The probability map is binarized with a specified threshold, and connected components are extracted from the binary map. As stated in a previous work~\cite{liao2020real}, the size of the extracted regions is smaller than the actual text regions. Therefore, the extracted regions are dilated using the Vatti clipping algorithm with offset \(D\), \(D = \frac{A\times r}{L}\),
where \(A\) is the area of a polygon region, \(L\) is the perimeter of a polygon, and \(r\) is the pre-defined dilation factor. After extracting the polygon from each dilated region, we calculate the center coordinates and give it as the reference point to the decoder. Finally, the decoder greedily predicts character sequences in the corresponding text region.

\section{Experiments}

\begin{table*}
\centering
\begin{tabular}{lcccccccccccccc}
\toprule
\multirow{2}{*}{Method}   & \multicolumn{3}{c}{Detection}    & \multicolumn{4}{c}{E2E} \\
                          & Recall    & Precision & F-measure & Strong    & Weak      & Generic & None\\
\midrule
TextNet* \cite{sun2018textnet}            & 85.41     & 89.42     & 87.37     & 78.66     & 74.90     & 60.45 & - \\
E2E TextSpotter \cite{he2018end}      & 86.00     & 87.00     & 87.00     & 82.00     & 77.00     & 63.00 & - \\
MaskTextSpotter* \cite{lyu2018mask}     & 81.00     & 91.60     & 86.00     & 79.30     & 73.00     & 62.40 & - \\
FOTS* \cite{liu2018fots}                 & 87.92	  & 91.85	  & 89.84	  & 83.55	  & 79.11	  & 65.33 & - \\
TextDragon \cite{feng2019textdragon}         & 83.75     & 92.45     & 87.88     & 82.54     & 78.34     & 65.15 & - \\
CharNet* \cite{xing2019convolutional}             & \textbf{90.47}     & \underline{92.65}     & \underline{91.55}     & 85.05     & 81.25     & 71.08 & 67.24 \\
Qin \etal \cite{qin2019towards}            & \underline{87.96}	  & 91.67	  & 89.78	  & \textbf{85.51}	  & \underline{81.91}	  & 69.94 & - \\
CRAFTS \cite{baek2020character}               & 85.30     & 89.00     & 87.10    & 83.10     & \textbf{82.10}     & 74.90 & \textbf{74.90} \\
MaskTextSpotter v3 \cite{liao2020mask}    & -         & -         & -         & 83.30     & 78.10     & 74.20 & - \\
Boundary \cite{wang2020all}             & 87.50     & 89.80     & 88.60     & 79.70     & 75.20     & 64.10 & - \\
TextPerceptron \cite{qiao2020text}       & 82.50     & 92.30     & 87.10     & 80.50     & 76.60     & 65.10 & - \\
ABCNet v2* \cite{liu2020abcnet}           & 86.00     & 90.40     & 88.10     & 83.00     & 80.70     & \underline{75.00} & - \\
MANGO \cite{qiao2020mango}                & -         & -         & -         & \underline{85.40}     & 80.10     & 73.90 & - \\
PGNetPGNet \cite{wang2021pgnet}             & 84.80     & 91.80     & 88.20     & 83.30     & 78.30     & 63.50 & - \\
Li \etal \cite{wang2021towards}           & -         & -         & \textbf{91.60}     & 84.23	  & 78.04	  & 65.53 & - \\
\midrule
\textbf{Ours}               & 86.23     & \textbf{93.72}     & 89.82   & 82.71     & 79.07    & \textbf{75.64} & \underline{71.72} \\
\bottomrule
\end{tabular}
\caption{
Experiment results on IC15. Models with * use multi-scale inference, and strong/weak/generic columns mean recognition with strong, weak, and generic lexicons, respectively.
} 
\label{tab:exp_ic15}
\end{table*}

\begin{table*}
\centering
\begin{tabular}{lcccccc}
\toprule
\multirow{2}{*}{Method}  & \multicolumn{3}{c}{Detection}    & \multicolumn{2}{c}{E2E} \\
                         & Recall    & Precision & F-measure & None      & Full \\
\midrule
TextNet \cite{sun2018textnet}                  & 59.45     & 68.21     & 63.53     & 54.02     & -     \\
Mask TextSpotter \cite{lyu2018mask}        & 55.00     & 69.00     & 61.30     & 52.90     & 71.80 \\
TextDragon \cite{feng2019textdragon}              & 75.70     & 85.60     & 80.30     & 48.80     & 74.80 \\
CharNet* \cite{xing2019convolutional}               & \underline{85.00}     & 88.00     & 86.50     & 69.20     & -     \\
PAN++ \cite{wang2021pan++}                   & 81.00     & 89.90     & 85.30     & 68.60     & 78.60 \\
Qin~\etal \cite{qin2019towards}               & \underline{85.00}     & 87.80     & 86.40     & 70.70     & -     \\
CRAFTS \cite{baek2020character}                  & \textbf{85.40}     & 89.50     & \textbf{87.40}     & \textbf{78.70}     & -    \\
Mask TextSpotter v3 \cite{liao2020mask}     & -         & -         & -         & 71.20     & 78.40 \\
ABCNet v2* \cite{liu2020abcnet}              & 84.10     & \underline{90.20}     & \underline{87.00}     & 73.50     & 80.70 \\
MANGO \cite{qiao2020mango}                     & -         & -         & -         & 72.90     & \textbf{83.60} \\
Li~\etal* \cite{wang2021towards}             & 59.38     & 63.25     & 61.25     & 58.56     & -     \\
\midrule
\textbf{Ours}            & 81.44         & \textbf{90.44}         & 85.71         & \underline{74.84}         & \underline{81.34}    \\
\bottomrule
\end{tabular}
\caption{
Experiment results on TotalText. Models with * use multi-scale inference. E2E result with Full and None types mean that recognition is done with/without a lexicon, respectively.}
%
\label{tab:exp_tt}
\end{table*}

\begin{figure*}[t]
  \centering
    \includegraphics[width=0.24\linewidth]{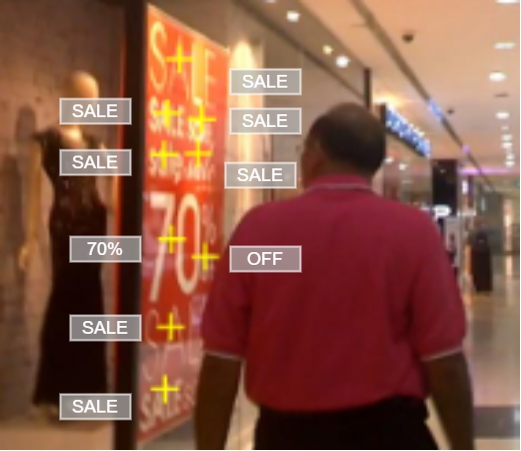}
    \includegraphics[width=0.24\linewidth]{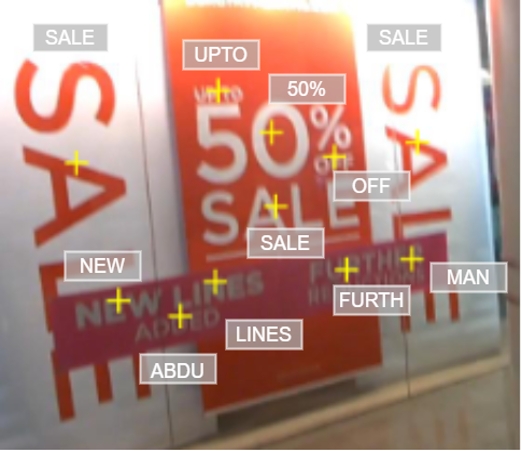}
    \includegraphics[width=0.24\linewidth]{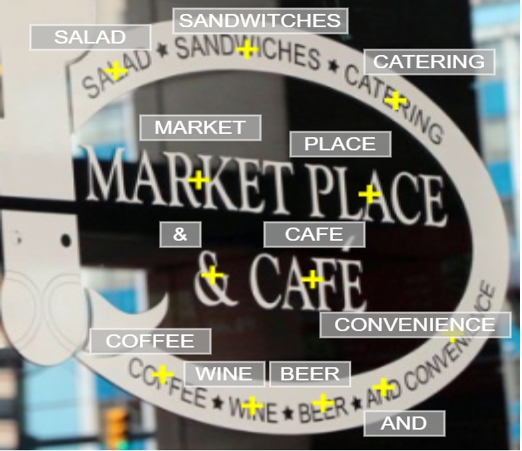}
    \includegraphics[width=0.24\linewidth]{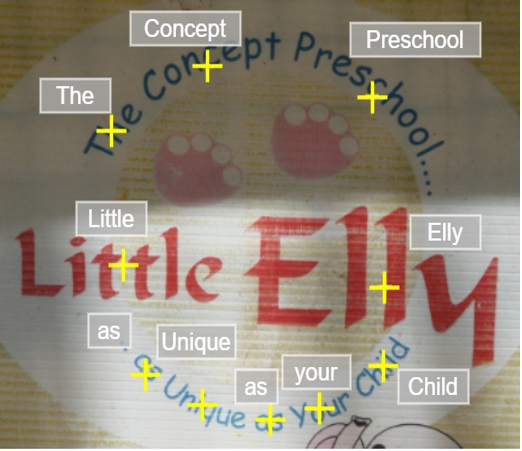}  
    \includegraphics[width=0.24\linewidth]{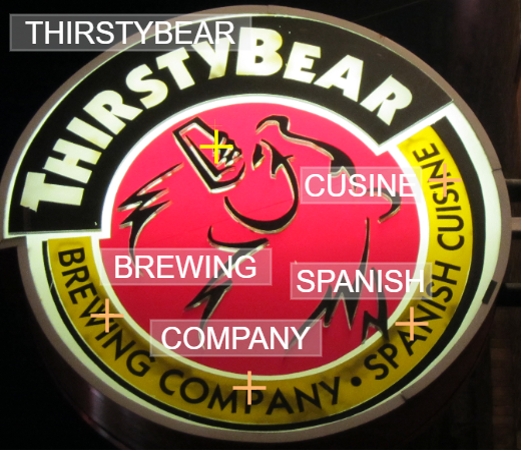}
    \includegraphics[width=0.24\linewidth]{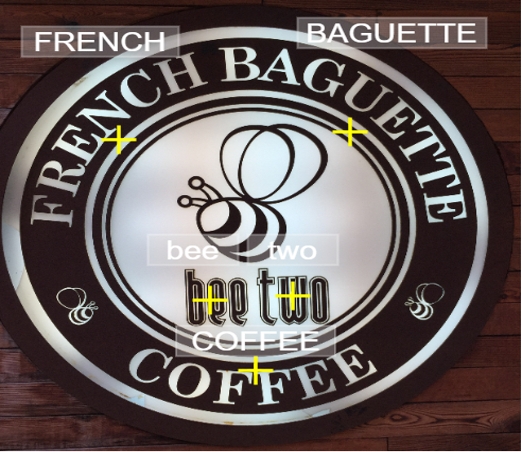}
    \includegraphics[width=0.24\linewidth]{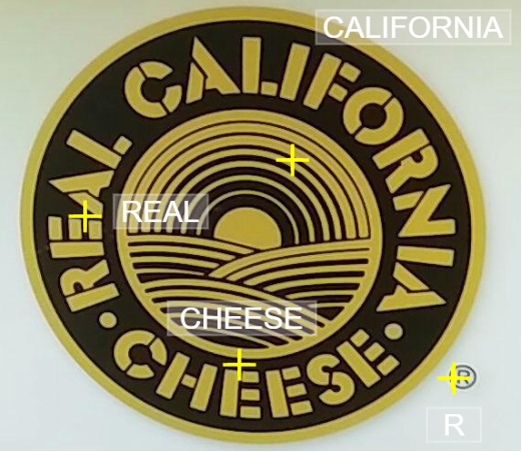}
    \includegraphics[width=0.24\linewidth]{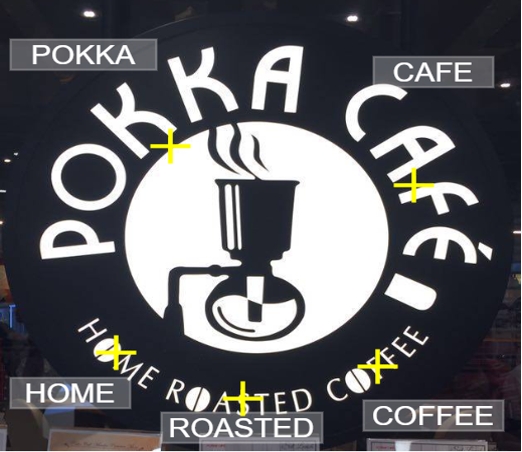}  
   \caption{Qualitative results of DEER on IC15~\cite{IC15} and TotalText~\cite{ch2017total} datasets. The reference point is presented as +. Please zoom in for better visualization.}
    \label{fig:qualitative_output}
\end{figure*}

\begin{figure*}[t]
  \centering
  \begin{subfigure}{0.49\linewidth}
    \includegraphics[width=0.32\linewidth]{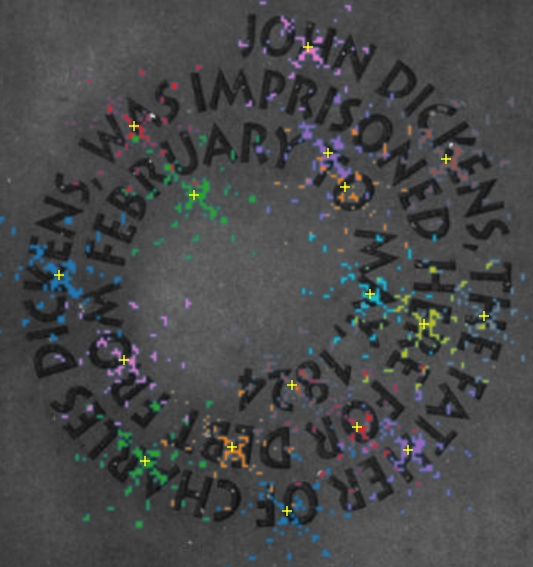}
    \includegraphics[width=0.32\linewidth]{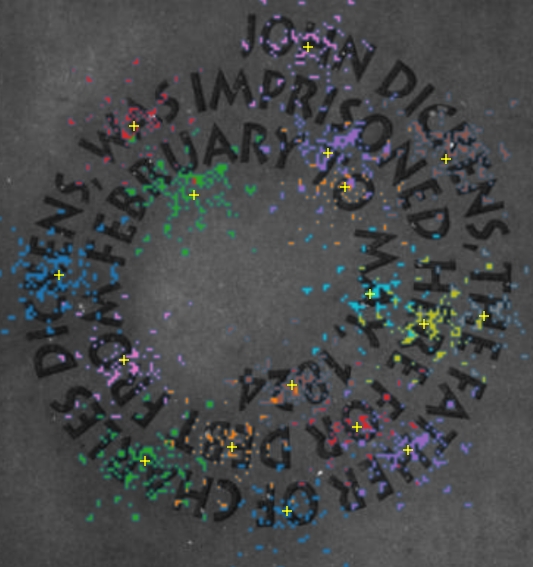}
    \includegraphics[width=0.32\linewidth]{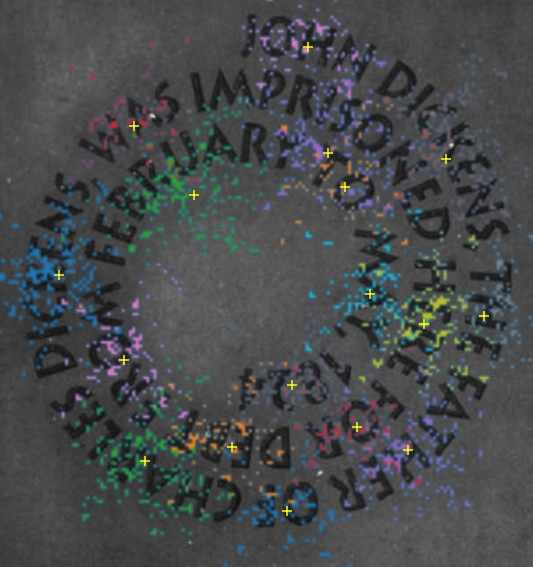}\\
    \caption{Visualized deformable attention map.}
    \label{fig:deformable_attention}
  \end{subfigure}
  \begin{subfigure}{0.49\linewidth}
    \includegraphics[width=0.32\linewidth]{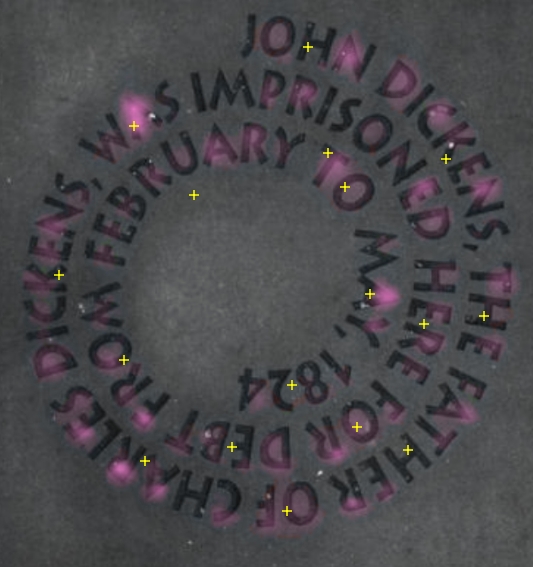}
    \includegraphics[width=0.32\linewidth]{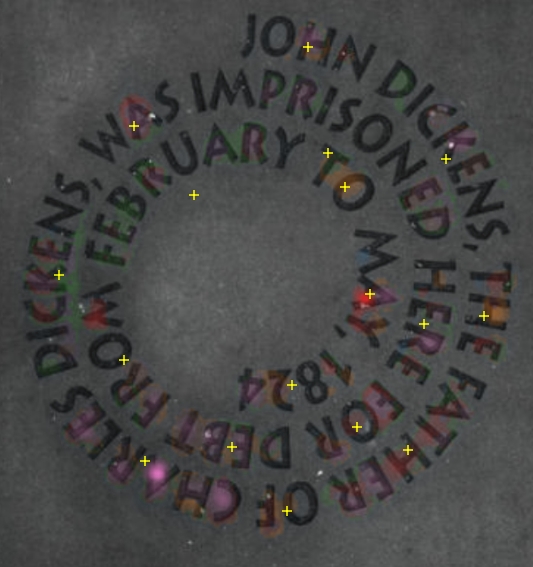}
    \includegraphics[width=0.32\linewidth]{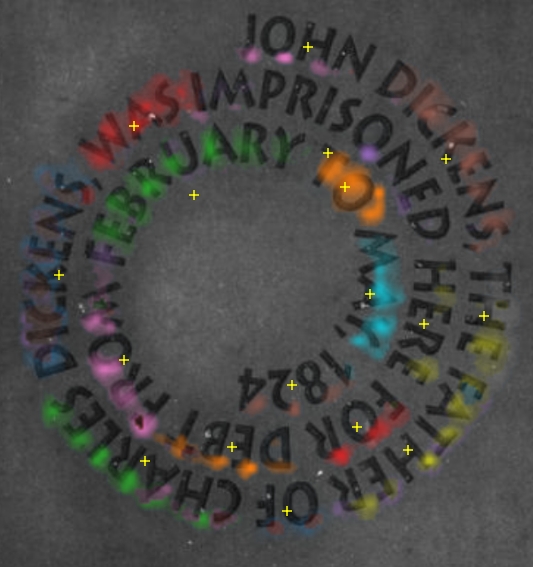}\\
    \caption{Visualized plain attention map.}
    \label{fig:plain_attention}
  \end{subfigure}
   \caption{Visualization of (a) deformable attention from layers 1, 3, and 5, (b) plain attentions from layers 2, 4, and 6, respectively. Each text instance is color-coded. }
    \label{fig:attentions}
\end{figure*}

\subsection{Datasets}

\noindent\textbf{TextOCR}~\cite{singh2021textocr} is a scene text dataset which contains arbitrary shaped text instances. It is annotated with polygons and contains 24,902 training images and 3,232 testing images.

\noindent\textbf{ICDAR 2015}~\cite{IC15} is a scene text dataset with quadrilateral text instances. It includes 1,000 training images and 500 testing images.

\noindent\textbf{TotalText}~\cite{ch2017total} is a scene text dataset with various text instances. It includes rotated and curved text instances, and is annotated using polygon bounding boxes. The dataset is composed of 1,255 training and 300 testing images.

\subsection{Implementation details}
The model is pretrained on TextOCR dataset for 400k steps with initial 10k warmup steps. We use Adam optimizer with a cosine learning rate scheduler. For pretraining, the learning rate is set to \(3e-4\), and the weight decay is set to \(1e-6\).  The pretrained model is then fine-tuned on each benchmark dataset for 10k steps with a learning rate of \(5e-5\). We use a batch size of 32 and sample 2 text instances on each image to train the recognizer.

For training data augmentation, we apply random rotation between -90 and 90 degrees, random resize between 50\%  and 300\%, safe random crop up to 640 pixels while preserving the bounding box, and color jitter with a probability of 0.8. For inference, the longer side of the input image is resized to 1280 or 1920 pixels while keeping the aspect ratio.

\subsection{Experimental Results}

On the ICDAR 2015 dataset, the end-to-end performance is evaluated under \textit{Strong}, \textit{Weak}, and \textit{Generic} contextualization settings. Each uses a separate lexicon to match the closest word produced by the model. Also, texts of length less than 3 are ignored, and special characters in the first or last position of a word are removed. The performance of \methodname~on ICDAR 2015 dataset is listed in \Table{exp_ic15}. With a score of 75.6, the model achieves state-of-the-art results under the generic contextualization setting. Note that the proposed method shows a lower detection score compared to previous works \cite{wang2021towards,xing2019convolutional}, but it achieves higher end-to-end performance. This result indicates that the recognition performance is more robust to detection error compared to previous works.

For the TotalText dataset, we measure the end-to-end performance in two ways. \textit{None} indicates that the model does not use any lexicon for evaluation, and \textit{Full} denotes that the model fully exploits a lexicon with the ground truth labels. \Table{exp_tt} shows the quantitative results. \methodname~achieves competitive results without feature pooling or masking.

For qualitative results, we show eight examples in \Figure{qualitative_output}. The yellow cross represents the reference point.
Using only a single point as a cue, \methodname~is able to correctly recognize texts in complex and dense scenes. Specifically, as shown in the first and second examples, we see that \methodname~successfully handles complex layouts such as text-in-text, crossing texts, rotated texts, and diverse font sizes.
\methodname~can also handle text instances in perspective and arbitrarily shaped texts. 
We also acknowledge that our text decoder does not highly depend on detected text regions. As can be seen in \Figure{teaser_mts_deer}, our model correctly predicts text sequence by attending to the extracted full image features even if detection results are inaccurate. We plan to present more visual results as supplementary material.

We visualize the attention map in \Figure{attentions} to analyze the roles of deformable attention and plain attentions. 
\Figure{deformable_attention} shows that deformable attention attends to the feature around the reference points in the first layer. As the layer progresses, it attends to the features around the entire text instances.
On the other hand, plain attention in \Figure{plain_attention} attends to the entire text instances in the scene at the first layer. For this reason,  the attention maps of each text instance are overlapped. In higher layers, plain attention attends to each character in the text specified by the reference point.

\begin{table}
\centering
\begin{tabular}{@{}lccc@{}}
\toprule
Dataset &  Perturbation & Detection & E2E \\
\midrule
\multirow{2}{*}{ICDAR 2015} & \checkmark & 89.82  & 71.72     \\
 & & 89.11 & 70.63    \\
\midrule
\multirow{2}{*}{TotalText} & \checkmark & 85.71  & 75.32   \\
 &    & 84.19  & 74.91    \\
\bottomrule
\end{tabular}
\caption{
\textbf{Pertubation} -- Applying perturbation on reference point \(p_\text{ref}\) during training is beneficial for the model performance, and it suggests that the model could be trained with noisy supervisions.
} 
\label{tab:perturb}
\end{table}

\begin{figure}[t]
  \centering
    \includegraphics[width=1.0\linewidth]{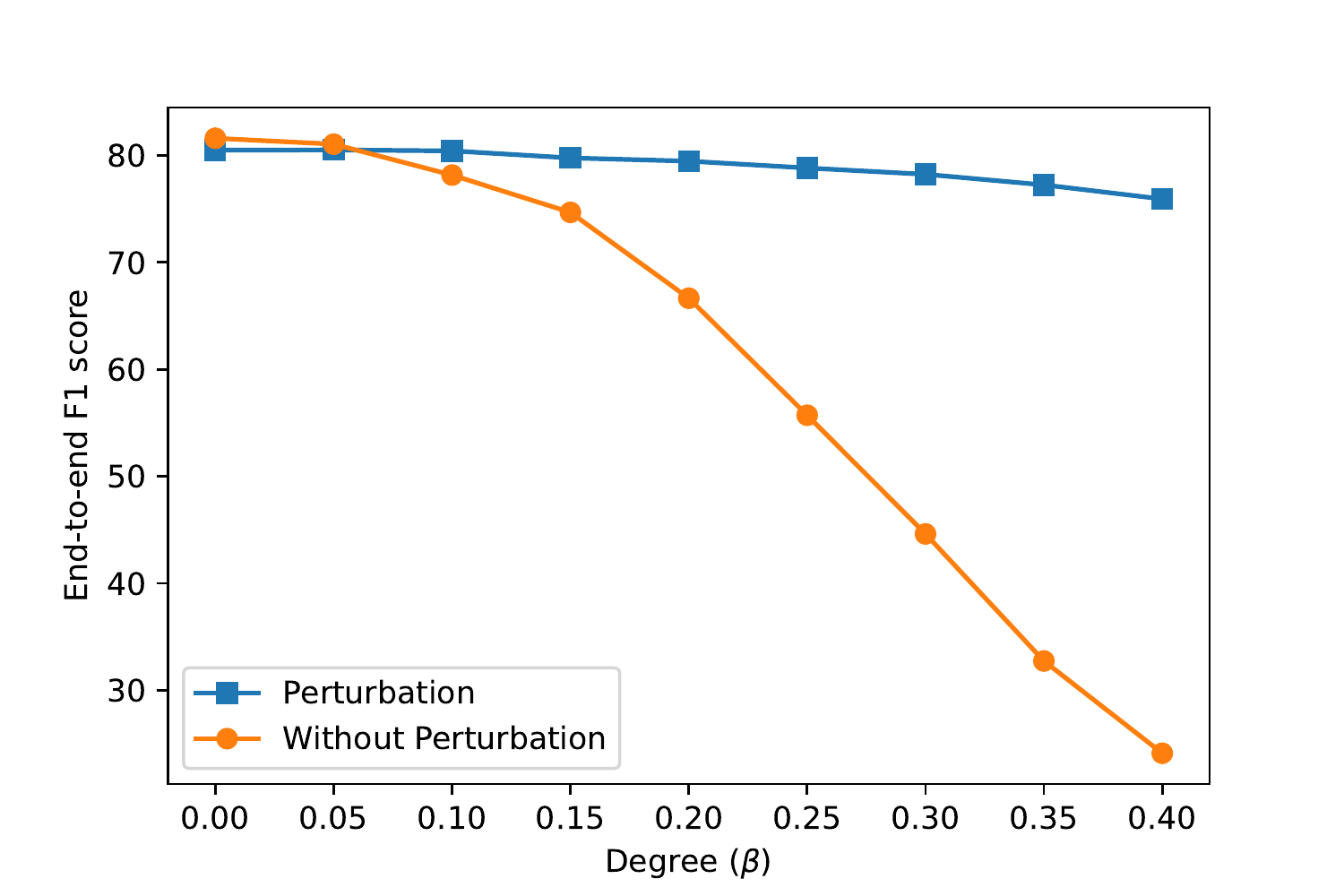}
   \caption{Moving the reference point during inference from the center point of the bounding boxes to their top-left points by degree. A model trained with perturbation is robust to noise caused by the moving of reference points.}
    \label{fig:move-point}
\end{figure}


\begin{table}
\centering
\begin{tabular}{@{}lccc@{}}
\toprule
Dataset & Sampling & Detection & E2E \\
\midrule
\multirow{2}{*}{ICDAR 2015} & Center & 89.82 & 71.72     \\
 & Inner  & 88.58 & 71.01    \\
 \midrule
\multirow{2}{*}{TotalText}  & Center & 85.71 & 75.32   \\
 & Inner    & 84.19 & 75.47    \\
\bottomrule
\end{tabular}
\caption{
\textbf{Center point \vs Inner point} -- The proposed method could be trained to use a more divers set of points inside the text polygon. This suggest that the model could recognize text sequences correctly even when the reference point is located outside of the text instance.
} 
\label{tab:alternative}
\end{table}

\begin{table}
\centering
\resizebox{\linewidth}{!}{ 
\begin{tabular}{@{}lcccc@{}}
\toprule
                        & Detection & Recall & Precision & F1 \\
\midrule
\multirow{2}{*}{ICDAR 2015} & \checkmark & 74.39     & 75.22     & 74.80    \\
 &             & 72.85     & 73.55     & 73.20    \\
\midrule
\multirow{2}{*}{TotalText} & \checkmark & 80.19    & 80.81        & 80.50   \\
 &  & 78.87    & 76.91     & 79.24    \\
\bottomrule
\end{tabular}
} 
\caption{
\textbf{Detection Supervision} -- Training the model to detect text instances is beneficial for the recognition of the text sequences.
} 
\label{tab:detect}
\end{table}



\subsection{Ablation study}

To analyze whether our proposed method is robust to the detection errors, we conduct ablation experiments on ICDAR 2015 and TotalText datasets. 
No lexicon is used throughout these experiments.



\noindent\textbf{Effectiveness of reference point perturbation} 
\Table{perturb} shows \methodname~trained with or without the perturbation of the center coordinates \(p_\text{ref}\) using \Equation{perturb}. The model trained with perturbation shows slightly better performance on both datasets.

\noindent\textbf{Robustness against detection error}
When detection results are not accurate, it is not guaranteed that the reference point \(p_\text{ref}\) will be placed at the center point of the polygon. To simulate this error, we deliberately move the ground truth reference point from the center point towards the top-left point by \(\beta\). Therefore, the new center point is \(p'_\text{ref} = (1 - \beta) p_\text{ref} + \beta p_\text{tl}\), where \(p_\text{ref}\) is the original center point and \(p_\text{tl}\) is the top-left point of the bounding polygon.


\Figure{move-point} shows the end-to-end F1-score on TotalText when \(\beta\) changes from 0.0 to 0.4. While the noisy reference points deteriorate the overall performance, the proposed model trained with reference point perturbation is much more robust to detection errors than the one trained without it.

\noindent\textbf{Alternative selection of reference point} 
For a text instance with extreme shape, its reference point obtained by simply taking the average of the polygon vertices might be located outside the bounding polygon.
In this experiment, we apply an alternative reference point sampling method, namely the \textit{Inner} method. For training, we simply use a random reference point inside the annotation polygon. For inference, we take the midpoint of the center cross section of the polygon.
Robustness against perturbing the reference points and alternative strategy for the sampling shows that \methodname~could be trained with noisy annotations.

\Table{alternative} compares the performance of the proposed model using the Inner method to the original one. The original sampling method performs comparably to the Inner method. This result indicates that reference points do not have to necessarily be located inside the text area for correct recognition; therefore, the proposed method is robust to recognizing text instances with extreme shapes.

\noindent\textbf{Training with detection supervision}
The proposed model does not directly use the region information of the text instances except for the reference points. However, the detection could guide the features to be beneficial for the recognizer. 
To verify this idea, we train \methodname~without detection loss. The model under this setting does not produce any detection results. Therefore, during evaluation, we use ground truth annotations to obtain the reference points.

\Table{detect} shows that the detection supervision benefits the recognizer. However, we expect that using extended training and a larger dataset could reduce the gap, as detection supervision is only used for guidance during training. We also believe that high-performing end-to-end text spotting models can be trained through just point supervision, eliminating the need for expensive polygon annotations.

\section{Discussions}
\begin{figure}[t]
  \centering
    \includegraphics[width=0.49\linewidth]{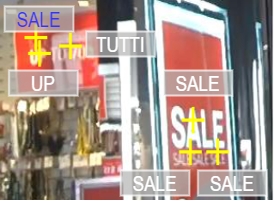}
    \includegraphics[width=0.49\linewidth]{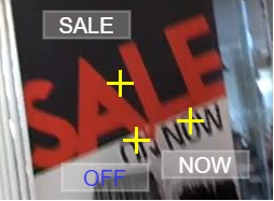}
   \caption{Structured error patterns in IC15. In the left image, \methodname~predicts `SALE' on ambiguous texts with red backgrounds as the word `SALE' frequently occurs on red backgrounds. On the right, as word `off' frequently occurs with the word `SALE', `OFF' is predicted instead of `ON,' which is near the word `SALE'.}
    \label{fig:potential}
\end{figure}

Allowing the recognizer to utilize the entire feature map mitigates the need for accurate detection of text instances, but it could be inefficient when the detection result contains the text instances neatly and the text is unambiguous.

Although it is not required to detect regional information of the text instances, it is required to predict it for the evaluation used in end-to-end text spotting metrics. We suspect that evaluation metrics with relaxed requirements on detecting precise regions could allow interesting research directions.

As the recognizer crucially depends on the detection of reference points and the recognition of each text instance is independent, it is not possible to detect missing text instances or duplicated reference points referring to the same instances.

Using surrounding features that are not restricted within text boundaries and utilizing contextual information could be beneficial to recognize ambiguous characters. However, it could enable the model to utilize spurious cues for recognition. It could make the model have structured error patterns in the recognition results, as shown in \Figure{potential}. Thus it is possible for the model to amplify the bias in the trained datasets.

\section{Conclusions}
We propose \methodname, detection-agonistic end-to-end recognizer for scene text spotting. By relaxing the coupling between the detection of the text bounding polygon and the recognition of text instances, the proposed model allows for the recognition of arbitrary-shaped text instances without the need for rather expensive polygon annotations and elaborate pooling mechanisms. Our experiments show that it is possible to have competitive performances on scene text spotting without using polygon proposals and pooling operations. We hope that the proposed detection-agnostic architecture could help construct scene text spotting models that are more flexible and easily trained, and offer insights to other vision tasks that require the generation of structured outputs.

{\small
\bibliographystyle{ieee_fullname}
\bibliography{cvpr}
}

\end{document}